%% file: root.tex
\documentclass[letterpaper, 10 pt, conference]{ieeeconf}
\IEEEoverridecommandlockouts    
\makeatletter
\def\endthebibliography{%
  \def\@noitemerr{\@latex@warning{Empty `thebibliography' environment}}%
  \endlist
}
\makeatother

\usepackage{bm}
\usepackage{cite}
\usepackage{array}
\usepackage{xcolor}
\usepackage{amsmath}
\usepackage{amssymb}
\usepackage{graphicx}
\usepackage{algorithm}
\usepackage{subcaption}
\usepackage{algorithmicx}
\usepackage{algpseudocode}
\usepackage[utf8]{inputenc}
\usepackage[dvipsnames]{xcolor}  
\usepackage[bookmarksopen, bookmarksnumbered, colorlinks=true, linkcolor=black, urlcolor=BlueViolet, citecolor=black,]{hyperref}

\usepackage{tikz}
\usepackage{pgfplots}
\pgfplotsset{compat=newest}
\usepgfplotslibrary{groupplots}

\input{./tools/paper_macros}
\usepackage[nolist]{acronym}
\input{./tools/acronyms}

\allowdisplaybreaks

\usepackage{setspace}
\newcommand{\reducespace}{\vspace{-0.3cm}} 

\title{\textbf{SM$^2$ITH}: \textbf{S}afe \textbf{M}obile \textbf{M}anipulation with \textbf{I}nteractive Human Prediction via \textbf{T}ask-\textbf{H}ierarchical Bilevel Model Predictive Control}

\author{
    Francesco D'Orazio$^{*1}$, Sepehr Samavi$^{*2,3}$, Xintong Du$^{*2}$, Siqi Zhou$^{3,4}$, Giuseppe Oriolo$^1$, Angela P. Schoellig$^3$
    \thanks{$^*$equal contribution.}
    \thanks{$^1$Department of Computer, Control and Management Engineering, of Sapienza University of Rome, Italy.}
    \thanks{$^{2}$University of Toronto Institute for Aerospace Studies (UTIAS) and the Vector Institute for Artificial Intelligence, Canada.}%
    \thanks{$^3$Learning Systems and Robotics lab at the Technical University of Munich and the Munich Institute for Robotics and Machine Intelligence (MIRMI), Germany.}
    \thanks{$^4$School of Computing Science, Faculty of Applied Sciences, Simon Fraser University, Burnaby, BC, Canada.}
    \thanks{Emails: \tt\small{\{dorazio, oriolo\}@diag.uniroma1.it, \{sepehr, xintong.du\}@robotics.utias.utoronto.ca, angela.schoellig@tum.de, siqi@sfu.ca}.
    }
}

\begin{document}

\maketitle

\begin{abstract}
    Mobile manipulators are designed to perform complex sequences of navigation and manipulation tasks in human-centered environments. While recent optimization-based methods such as \ac{HTMPC} enable efficient multitask execution with strict task priorities, they have so far been applied mainly to static or structured scenarios. Extending these approaches to dynamic human-centered environments requires predictive models that capture how humans react to the actions of the robot. This work introduces \ac{\smith}, a unified framework that combines \ac{HTMPC} with interactive human motion prediction through bilevel optimization that jointly accounts for robot and human dynamics. The framework is validated on two different mobile manipulators, the Stretch~3 and the Ridgeback–UR10, across three experimental settings: \textit{(i)} delivery tasks with different navigation and manipulation priorities, \textit{(ii)} sequential pick-and-place tasks with different human motion prediction models, and \textit{(iii)} interactions involving adversarial human behavior. Our results highlight how interactive prediction enables safe and efficient coordination, outperforming baselines that rely on weighted objectives or open-loop human models. Code: \url{https://github.com/learnsyslab/sm2ith.git}
\end{abstract}
\acresetall

\section{Introduction}

Mobile manipulators, which combine a mobile base with one or more robotic arms, are envisioned as versatile service robots capable of performing complex tasks in human-centered environments. For example, a robot waiter may need to retrieve a tray of dishes from the kitchen, transport them across a crowded dining hall while ensuring no spillage, and deliver them to the right guests. Such scenarios reveal two interconnected challenges. First, the robot needs to seamlessly coordinate multiple manipulation and navigation tasks with different priorities. Second, the robot needs to simultaneously negotiate safe (i.e., collision-free) and efficient navigation among a crowd of humans.

Towards addressing the first challenge, early approaches have been specialized toward specific notions of prioritization. To execute a temporal sequence of tasks (e.g., retrieve dish before going to guest) some methods propose decoupling base and arm planning to execute one task at a time~\cite{carius2018deployment}, which can lead to slow execution as the robot fails to take advantage of kinematic redundancy. To simultaneously execute tasks of varying importance (e.g., avoid spilling even at the expense of navigation progress), other methods propose using weighted sums of task costs~\cite{sathya2021weighted}, which relies on precise tuning to handle tradeoffs between contradictory objectives.
However, in such redundant systems, prioritization can be formulated as a strict hierarchy. Thus, recent methods, such as \ac{HTMPC}~\cite{du2023hierarchical}, propose using lexicographic optimization~\cite{ehrgott2005multicriteria}, where the highest priority task is first optimized in isolation and lower priority tasks are consecutively included without jeopardizing higher priority tasks. Yet, the focus of the aforementioned methods has not been on human-centered environments where safe navigation among humans is necessary.

\begin{figure}[t]
    \centering
    \includegraphics[width=0.99\columnwidth]{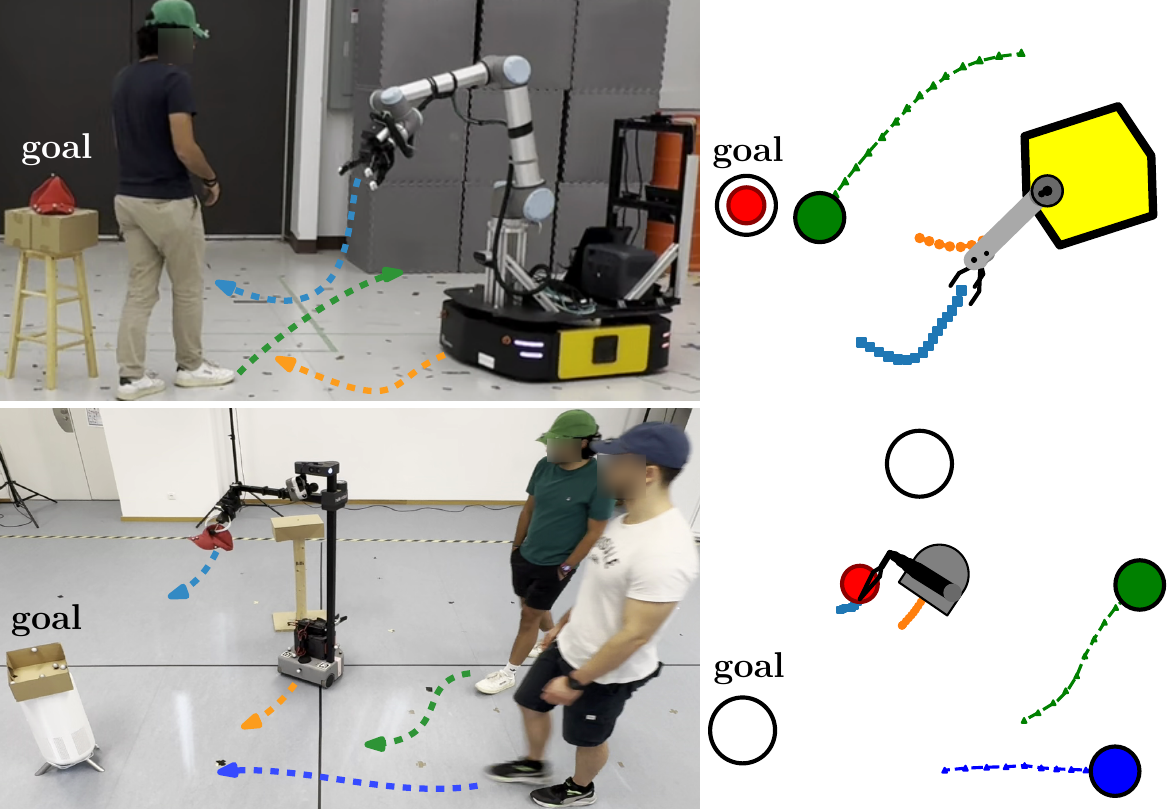}
    \caption{Snapshots of pick-and-place tasks in human-centered environments with the Ridgeback–UR10 (top) and Stretch~3 (bottom) platforms along with a top-down view animation. The illustrated \smith{} solutions consist of planned robot trajectories coupled with human predictions allowing the robot to interactively navigate among humans while coordinating multiple prioritized mobile manipulation tasks. Video: \url{http://tiny.cc/sm2ith}.}
    \label{fig:front_figure}
    \reducespace
\end{figure}

To address the second challenge, the robot must anticipate human movement and adjust its trajectory in real-time. Some methods propose to reactively avoid dynamic obstacles by tracking occupancy grids over time \cite{burgess2024reactive} or using a geometric local motion planner \cite{merva2025globally} to avoid humans. While these methods show promise in executing collision-free mobile manipulation tasks, they neglect the fact that the motions of the humans and the robot influence each other. As such, a variety of work has focused on generating human trajectory predictions in open-loop (e.g., \cite{gu2022mid}) and using them as an input in the controller \cite{poddar2023fromcrowdmotiontorobot}. However, this approach has been shown to lead to overly conservative or even frozen behaviors in crowded scenes \cite{trautman2010unfreezing}. Thus, more recent approaches propose to model the interaction between the humans and the robot by integrating the prediction model into the robot controller and incorporating collision avoidance as a cost term \cite{schaefer2021leveraging} or as constraints \cite{samavi2024sicnav}. However, to the best of the authors' knowledge, such interactive formulations have not been used in mobile manipulation, only mobile navigation.

In this work, we propose \ac{\smith}, a joint prediction and control framework that simultaneously predicts human trajectories and generates robot commands. Our approach builds on \ac{HTMPC} \cite{du2023hierarchical} and \ac{SICNav} \cite{samavi2024sicnav} to enable a mobile manipulator to execute multiple prioritized base and manipulation tasks \cite{du2023hierarchical} while reasoning about, and accounting for, human responses in the shared space. \ac{\smith} models humans to be solving the \ac{ORCA} \cite{van2011reciprocal} optimization problem, and embeds this model as constraints within \ac{HTMPC} resulting in a bilevel problem \cite{samavi2024sicnav}. This way, the predictions used for robot planning are closed-loop: our algorithm couples robot commands and associated human predictions by assuming that the humans solve \ac{ORCA} in reaction to robot motion. To ensure that robot motions remain in the set of safe collision-free states, we introduce \acp{CBF} \cite{ames2019control} that limit how quickly the robot approaches unsafe states. Fig.~\ref{fig:front_figure} illustrates the jointly planned trajectory for a pick-and-place application using two different mobile manipulator platforms.

Our contributions are twofold. \underline{First}, we formulate \ac{\smith}, a hierarchical task  mobile manipulation control framework with interactive human predictions. Specifically, we ($i$) extend \ac{SICNav}'s collision avoidance constraints to a 3D setting for whole-body (base and arm) avoidance, ($ii$) integrate \ac{SICNav}'s bilevel structure into \ac{HTMPC}'s lexicographic formulation to simultaneously predict human motion while solving for robot actions, and ($iii$) introduce \acp{CBF} to proactively satisfy safety constraints (i.e., collision avoidance) in human-centered environments. \underline{Second}, we conduct extensive experiments (80 runs) to evaluate the performance of \ac{\smith} compared to methods using baseline models of human motion in sequential pick-and-place tasks using two different mobile manipulator platforms illustrated in Fig.~\ref{fig:front_figure}. We also conduct further experiments~(60 runs) to evaluate the performance of \ac{\smith} against a weighted-sum baseline controller on simultaneous mobile manipulation tasks with different priorities.
To the best of the authors' knowledge, this is the first whole-body control method for mobile manipulators in human-centered environments that couples human motion predictions with robot actions to explicitly model interactions during collision avoidance.

\section{Related Work}
\subsection{Mobile Manipulation with Task Prioritization}
Early motion planning methods for mobile manipulation relied on decoupled task-space planners for the base and end-effector, executed sequentially \cite{carius2018deployment} or through loosely coordinated controllers \cite{limerick2023architecture}. While reactive, these approaches treated tasks in isolation and could not leverage redundancy effectively. Weighted-sum formulations \cite{sathya2021weighted} allowed simultaneous execution but requires fine tuning to trade off contradictory objectives, while whole-body joint-space planners \cite{zimmermann2021go, thakar2020manipulator} improved coordination at the expense of heavy computation and limited reactivity.

Hierarchical-task control resolves these issues by enforcing strict priorities, first through inverse kinematics and null-space methods \cite{hanafusa1981analysis} and later through hierarchical quadratic programming \cite{escande2014hierarchical, wang2024hierarchical, nie2022hqp}. Embedding such hierarchies within predictive control has led to \ac{HTMPC}, which enforces lexicographic priorities over a horizon and has been validated on mobile manipulators \cite{du2023hierarchical} and redundant arms \cite{wang2024himpc}.

While these methods have demonstrated efficient multitask execution, their applications are focused on static or structured settings. Mobile manipulation in human-centered environments remains an open challenge: here, the robot must not only resolve multiple objectives but also account for the fact that its own behavior can affect human motion, and vice versa. This motivates the integration of \ac{HTMPC} with predictive models that capture human–robot interaction.

\subsection{Dynamic Obstacle and Human Avoidance}

One straightforward strategy for human avoidance is to model them as dynamic obstacles and leverage standard safe control frameworks to enforce collision avoidance constraints. For instance, dynamic obstacle avoidance has been integrated into predictive control through whole-body \ac{MPC} with signed-distance constraints \cite{heins2023keep, tarantos2024navigation, chiu2022collision, gaertner2021collision}, perceptive \ac{MPC} for high \ac{DOF} robots \cite{pankert2020perceptive}, and \acp{CBF} for safety filtering \cite{bena2025geometry, xu2024dynamic, jian2023dynamic}. Uncertainty-aware approaches extend these methods via chance constraints \cite{de2023scenario}, and occlusion-aware \ac{MPC} \cite{roya2025oampc}. When extending these methods for humans in shared spaces, it is often the case that human motion is predicted in an open-loop manner (e.g., constant velocity)~\cite{vulcano2022safe}.

For human interaction, however, simple predictors combined with collision constraints are insufficient. In particular, the absence of explicit modeling of interaction (i.e., the fact that humans and robots adapt to each other's behavior in shared space) can result in deadlocks or conservative behaviors. To address this issue, richer models such as the social force model \cite{jafari2024time} and \ac{RVO} frameworks \cite{van2008reciprocal} have been proposed to capture the reciprocal adaptation behavior. Moreover, recent extensions formulate joint human–robot prediction as optimal control problems \cite{van2011reciprocal, samavi2024sicnav}. These advances improve realism in navigation but remain focused on mobile robots without arms and base-level motion. Integration with multitask, redundancy-aware control for mobile manipulators is still underexplored. This work addresses that gap by proposing the \ac{\smith} framework, embedding interactive prediction within the \ac{HTMPC} formulation for mobile manipulation tasks.

\section{Problem Formulation}\label{sect:problem_formulation}

Consider a mobile manipulator operating in a three-dimensional workspace, $\workspace$, shared with humans. We denote the robot state as, $\robstate = (\robconfig, \robpseudovel)$, where $\robconfig \in \Reals^\configdim$ is the configuration vector that includes both the base variables~$\robconfigat{b}$ and arm variables~$\robconfigat{a}$. The pseudo-velocities, $\robpseudovel \in \Reals^\pseudoveldim$, are given by $\robvel = \nullkinmatatstate\robpseudovel$, where $\nullkinmat$ is a block diagonal matrix consisting of a state-dependent component for the base and an identity matrix for the arm variables \cite{siciliano2025foundations}. The input of the robot is the time derivative of the pseudo-velocity, $\control = \robpseudoacc$.

The workspace $\workspace$ contains $\nhum$ humans indexed by $\forhumin$ and $\nobs$ static obstacles indexed by $\forobsin$. We define a global system state that concatenates the robot state and the states of all humans, $\state = (\robstate, \id{1}\state, \dots, \id{\nhum}\state)$. Each human state is defined by the agent's Cartesian position and velocity, $\id{j}\state \in \Reals^4$, and human dynamics are modeled as a kinematic integrator.

The objective of our proposed controller is to generate real-time robot actions such that
\begin{itemize}
    \item[\textit{(i)}] collisions with humans and static obstacles are avoided at all times,
    \item[\textit{(ii)}] the motions remain kinematically feasible (i.e., within the kinodynamic limits of the robot), and
    \item[\textit{(iii)}] the robot executes multiple tasks for both base and end effector, with a strict prioritization, while minimizing violation of the task costs.
\end{itemize}

\section{Proposed Approach}

\begin{figure}
    \centering
    \includegraphics[width=\linewidth]{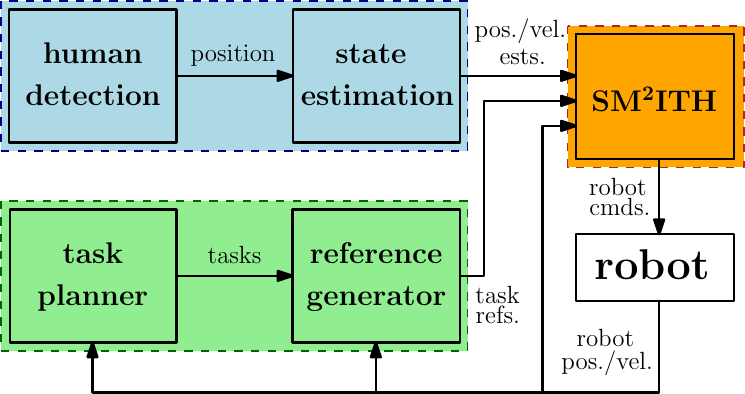}
    \caption{Proposed control architecture composed of three main blocks. The motion generation block (orange) computes the control commands using a hierarchical predictive formulation that accounts for both robot tasks and predicted human motion. The planning block (green) organizes the tasks into a priority hierarchy and generates the corresponding references to be tracked. The human estimation block (blue) detects humans in the environment and estimates their motion, providing predictions that feed into the control loop.}
    \label{fig:block_scheme}
    \reducespace
\end{figure}

To address the problem stated in Section \ref{sect:problem_formulation}, we propose the architecture illustrated in Fig.~\ref{fig:block_scheme}. The SM$^2$ITH block (orange), the primary contribution of this paper, is responsible for computing the robot commands accounting for future human trajectories. Our method builds upon the \ac{HTMPC} control scheme \cite{du2023hierarchical} allowing to solve prioritized tasks efficiently, which are defined as a lexicographic optimization problem \cite{ehrgott2005multicriteria}, and embeds an optimal model for human motion as lower-level constraints resulting in a bilevel MPC optimization problem \cite{samavi2024sicnav}.

The planning block (green in Fig.~\ref{fig:block_scheme}) provides the references that the robot must follow to execute the higher priority tasks at each time step. The human state estimation block (blue in Fig.~\ref{fig:block_scheme}) gives estimates of human positions and velocities through Kalman Filters.

\subsection{Hierarchical Task Model Predictive Control (HTMPC)}\label{sect:htmpc}

To solve a set of tasks ordered by strict prioritization, $\left[\taskcost_1(\state), \taskcost_2(\state), \dots, \taskcost_\ntasks(\state)\right]$, we use the \ac{HTMPC} framework~\cite{du2023hierarchical}. At each control step $\ctime$, the \ac{HTMPC} controller computes the optimal state and input trajectories $\mpcrobstate{}^*, \mpcrobcontrol{}^*$ over a finite time horizon $\horiz$ for a given sampling time $\deltat$, by solving the following lexicographic optimization problem:
\begin{subequations}\label{eq:htmpc}
    \begin{align}
        \lexmin_{\mpcrobstate, \mpcrobcontrol} & \ [\taskcost_1, \taskcost_2, \dots, \taskcost_\ntasks]
        \label{eq:htmpc_cost}\\
        \text{s.t. } & \mpcrobstateat{\kpone} = \robdyn(\mpcrobstateat{\dtime}, \mpcrobcontrolat{\dtime})
        \label{eq:htmpc_dyn}\\
        & \mpcrobstateat{\dtime} \in \sspace, \mpcrobcontrolat{\dtime} \in \uspace, \ \forall \dtime = 0, \dots, \horiz - 1
        \label{eq:htmpc_limits}\\
        & \mpcrobstateat{0} = \mpcrobstateat{\init}
        \label{eq:htmpc_init}\\
        & \text{\textbf{sd}}(\robstateat{\dtime}) \ge \scdist,
        \label{eq:htmpc_self_collision}
    \end{align}
\end{subequations}
where $\mpcrobstateat{\init}$ is the current robot state, and the prediction index $\dtime$ corresponds to time $\ctime + \dtime\deltat$. The optimization enforces robot dynamics \eqref{eq:htmpc_dyn}, state and input limits \eqref{eq:htmpc_limits}, initial condition \eqref{eq:htmpc_init}, and self-collision avoidance \eqref{eq:htmpc_self_collision} defined as a signed distance between robot links.
Each task cost, $\taskcost_\idtask$, represents the accumulated tracking error over the horizon, $\taskcost_\idtask = \frac{1}{2} \sum_{\dtime=0}^{\horiz-1} \norm{\err_{\idtask\dtime}}^2_{\costweight_\dtime} + \norm{\err_{\idtask\horiz}}^2_{\costweight_\horiz}$, with $\err \in \Reals^\idtaskdim$, $\norm{\err}^2_{\costweight} = \trans\err\costweight\err$ and $\costweight \in \Reals^{\idtaskdim \times \idtaskdim}_+$.

To solve \eqref{eq:htmpc}, we break down the problem into a sequence of \ac{STMPC} problems, one per task:
\begin{subequations}\label{eq:stmpc}
    \begin{align}
        \min_{\mpcrobstate, \mpcrobcontrol} & \ \taskcost_\idtask + \taskcost_{\text{eff}}
        \label{eq:stmpc_cost}\\
        \text{s.t. } & \eqref{eq:htmpc_dyn}, \eqref{eq:htmpc_limits},\eqref{eq:htmpc_init},\eqref{eq:htmpc_self_collision} \label{eq:stmpc_sys_consts}\\
        & |\errscalar_{\dtime \idtaskdimmid}| \le |\errscalar_{\dtime \idtaskdimmid}^{\idtaskmid^*}|,
        \label{eq:stmpc_lex_constr}\\
        & \forall \idtaskmid = 1, \dots, \idtask - 1, \ \idtaskdimmid = 1, \dots, \idtaskdim \nonumber.
    \end{align}
\end{subequations}
Each \ac{STMPC} optimizes a single task cost $\taskcost_\idtask$, regularized by an effort cost $\taskcost_{\text{eff}}$, subject to system constraints \eqref{eq:stmpc_sys_consts}. After finding the optimal solution to the first \ac{STMPC} problem containing only the highest priority task cost, consecutive \ac{STMPC} problems are additionally subject to a \textit{lexicographic optimality constraint} \eqref{eq:stmpc_lex_constr}. This constraint enforces that at each step~$\dtime$ in the horizon, the tracking error of already-solved higher priority tasks does not increase when optimizing the current lower priority task (details can be found in Section~IV.D of \cite{du2023hierarchical}). This process leads to an \ac{HTMPC} solution that leverages mobile manipulators' kinematic redundancy to optimize multiple tasks while enforcing priority.

\subsection{Safe and Interactive Crowd Navigation (SICNav)}\label{sect:sicnav}

To generate coupled human predictions while optimizing the robot plan, we follow \ac{SICNav} \cite{samavi2024sicnav} and incorporate the \ac{ORCA} \cite{van2011reciprocal} problems as constraints in the robot's \ac{HTMPC} to obtain a bilevel optimization problem:
\begin{subequations} \label{eq:sicnav}
	\begin{align}
		\min_{\mpcstate, \mpccontrol, \mpcaction} \
        & \taskcost(\mpcstate, \mpccontrol)
        \label{eq:sicnav_min}\\
		\centremathcell{\text{s.t.}}
        & \mpcrobstateat{0} = \mpcrobstateat{\init}
        \label{eq:sicnav_initcondconst}\\
        & \mpcrobcontrolat{\dtime} \in \uspace, \ \mpcrobstateat{\dtime} \in \sspace
        \label{eq:sicnav_actionconst}\\
        & \rob{\mpcstateat{\kpone}} = \robdyn(\mpcrobstateat{\dtime}, \mpcrobcontrolat{\dtime})
        \label{eq:sicnav_rob_dynconst}\\
        & \mpchumstateat{\kpone} = \humdyn(\mpchumstateat{\dtime}, \mpchumactionat{\dtime})
        \label{eq:sicnav_hum_dynconst}\\
        & \mpchumactionat{\dtime} \in \id{\idA}{\orcarlxsolnset}(\mpcstateat{\dtime}).
        \label{eq:sicnav_llorca}\\
        & \forall \dtime = 0, \dots, \horiz - 1, \ \forall \idA = 1, \dots, \nhum \nonumber
	\end{align}
\end{subequations}
The \textit{upper-level} cost \eqref{eq:sicnav_min} and constraints \eqref{eq:sicnav_initcondconst}-\eqref{eq:sicnav_rob_dynconst} optimize robot trajectories, and the \textit{lower-level} \ac{ORCA} constraints~\eqref{eq:sicnav_llorca} enforce that human actions are consistent with \ac{ORCA} solutions.
Here, $\mpcaction$ denotes the human velocity actions that are themselves the solution to ORCA optimization problem~\eqref{eq:relaxed_orca_acc_argmin}. These actions~$\mpcaction$ are used as human trajectory predictions, not as control inputs to the system.

At each time step $\ctime$, the \ac{ORCA}-based human prediction model is formulated as follows:
\begin{subequations} \label{eq:relaxed_orca_acc_argmin}
    \begin{align}
        \id{\idA}{\orcarlxsolnset}(\stateat{\ctime}) := \argmin_{\orcacontrol,\orcaslack} & \twonorm{{\orcacontrol} - {{\vel_\orcapref}}(\stateat{\ctime})}^2 + \pnorm{\orcaslackpenal}{\orcaslack}^2
        \label{eq:relaxed_orca_acc_argmin_cost}\\
        \text{s.t. } & \twonorm{\orcacontrol}^2 \le \orcacontrolscalar_{\max}^2
        \label{eq:relaxed_orca_acc_argmin_vel_max}\\
        & \twonorm{\orcacontrol - \at{\vel}{\ctime}}^2 \le \deltat \accscalar_{\max} - \orcaslackacc
        \label{eq:relaxed_orca_acc_argmin_acc_max}\\
        & \trans{\orcasupcolldyn\orcalinedir(\stateat{\ctime})}\orcacontrol \ge \orcasupcolldyn\orcaconstgeq(\stateat{\ctime}) - \orcaslackvel
        \label{eq:relaxed_orca_acc_argmin_agent_agent}\\
        & \trans{\orcasupcollstatic\orcalinedir(\stateat{\ctime})}\orcacontrol \ge \orcasupcollstatic\orcaconstgeq(\stateat{\ctime})
        \label{eq:relaxed_orca_acc_argmin_agent_static}\\
        & \orcaslack \ge 0
        \label{eq:relaxed_orca_acc_argmin_slack},
    \end{align}
\end{subequations}
where, $\orcacontrol \in \Reals^2$ is the optimization variable, $\vel_\orcapref\in \Reals^2$ is some given desired velocity (e.g. latest measured human velocity), \eqref{eq:relaxed_orca_acc_argmin_vel_max} is the maximum velocity constraint, and \eqref{eq:relaxed_orca_acc_argmin_agent_agent}-\eqref{eq:relaxed_orca_acc_argmin_agent_static} encode human–human and human-robot, and human–static obstacle collision avoidance, respectively. We introduce slack variables $\orcaslack = (\orcaslackacc, \orcaslackvel)$ to ensure there always exists a feasible solution, 
while penalizing constraint violation with $\pnorm{\orcaslackpenal}{\orcaslack}^2$ (details can be found in Section~III.B of \cite{samavi2024sicnav}). We also introduce acceleration constraints \eqref{eq:relaxed_orca_acc_argmin_acc_max} to obtain smooth trajectory predictions.

To solve the bilevel problem \eqref{eq:sicnav}, we reformulate the problem to a single-level optimization by replacing the lower-level ORCA problems with their \ac{KKT} conditions.

\subsection{SM$^2$ITH Formulation}

We unify the frameworks of Sections~\ref{sect:htmpc} and \ref{sect:sicnav} into a single control scheme 
to obtain \ac{\smith}: 
\begin{subequations}\label{eq:smith}
    \begin{align}
        \lexmin_{\mpcstate, \mpccontrol, \mpcaction, \mpcslack, \mpcmul} & \ [\taskcost_1(\mpcrobstate, \mpcrobcontrol), \dots, \taskcost_\ntasks(\mpcrobstate, \mpcrobcontrol)]
        \label{eq:smith_cost}\\
        \text{s.t. } & \mpcrobstateat{\kpone} = \robdyn(\mpcrobstateat{\dtime}, \mpcrobcontrolat{\dtime})
        \label{eq:smith_robdyn}\\
        & \mpcrobstateat{\dtime} \in \sspace, \mpcrobcontrolat{\dtime} \in \uspace
        \label{eq:smith_limits}\\
        & \mpcrobstateat{0} = \mpcrobstateat{\init}
        \label{eq:smith_init}\\
        & \text{\textbf{sd}}(\robstateat{\dtime}) \ge \scdist
        \label{eq:smith_self_collision}\\
        & \cbfvec(\stateat{\kpone}) \ge (1 - \cbfparam)\cbfvec(\stateat{\dtime})
        \label{eq:smith_avoid_coll}\\
        & \mpchumstateat{\kpone} = \humdyn(\mpchumstateat{\dtime}, \mpchumactionat{\dtime})
        \label{eq:smith_humdyn}\\
        & \grad{\mpchumactionat{\dtime}, \mpchumslack}{\lagfun}(\mpchumactionat{\dtime}, \mpchumslack, \mpchummul) = \bm 0
        \label{eq:smith_kkt_stationarity}\\
        & \trans{\lagmulvec} \orcaconst(\mpchumactionat{\dtime}, \mpchumslack) = 0
        \label{eq:smith_kkt_complementarity}\\
        & \orcaconst(\mpchumactionat{\dtime}, \mpchumslack) \le \bm 0
        \label{eq:smith_kkt_primal}\\
        & \mpchummul \ge \bm 0.
        \label{eq:smith_kkt_dual}\\
        & \forall \dtime = 0, \dots, \horiz - 1, \ \forall \idA = 1, \dots, \nhum \nonumber
    \end{align}
\end{subequations}
In this formulation, we extend \ac{HTMPC} \eqref{eq:htmpc} by incorporating the \ac{KKT} conditions of \ac{ORCA} \eqref{eq:relaxed_orca_acc_argmin} as constraints \eqref{eq:smith_kkt_stationarity}-\eqref{eq:smith_kkt_dual} to obtain predictions of the humans in the environment. These predictions are coupled with the robot plan being optimized.

To ensure human–robot and human-obstacle collision avoidance, we introduce a \ac{DT-CBF} constraint~\eqref{eq:smith_avoid_coll}, which maintains the state of the system within the safe set $\safeset$, which in our case is the set of all collision-free states. The \ac{DT-CBF} framework~\cite{agrawal2017discrete, zeng2021safety, dorazio2024maintaining} is formulated by defining the safe set as the superlevel set of a function, $\safeset = \{\state \in \sspace\suchthat\cbffun(\state) \ge 0\}$, where $\cbffun(\state):\sspace \rightarrow \Reals$ \cite{ames2019control}. The inequality, $\cbffun(\stateat{\kpone}) \ge (1 - \cbfparam)\cbffun(\stateat{\dtime})$, ensures that $\safeset$ is forward invariant (i.e., the future state evolutions remain within $\safeset$ for all initial state $\stateat{\init} \in \safeset$). The hyperparameter $\cbfparam \in [0, 1)$ bounds how quickly the system's state can approach the boundary of the safe set.

In our case, to ensure collision avoidance with respect to humans in the shared space, we model the volume occupied by humans as cylinders centered at their respective predicted state $\id{\idA}{\state_\dtime}$ and define a set of \acp{DT-CBF}, denoted by $\id{\idA}\cbfvec(\stateat{\dtime})$, with individual functions capturing collision-avoidance constraints with respect to the robot base and each arm link. The collision avoidance constraints with respect to static obstacles are defined in a similar manner. The constraint function~$\cbfvec(\stateat{\dtime})$ is a concatenation of the constraints capturing collision avoidance with respect to all humans and static obstacles in the environment.

Intuitively, \acp{DT-CBF} guarantee that the robot's distance to humans and static obstacles are always above a safety threshold by bounding how quickly the robot can approach humans and obstacles. As a result, the \ac{DT-CBF} constraints encourage more conservative robot behavior closer to the boundary of the safe set, providing an extra layer of safety. In practice, the hyperparameter~$\gamma$ can be further used to control the level of cautiousness in the robot's behavior.

\section{Experiments}

\begin{figure}[t]
    \centering
    \includegraphics[width=0.85\columnwidth]{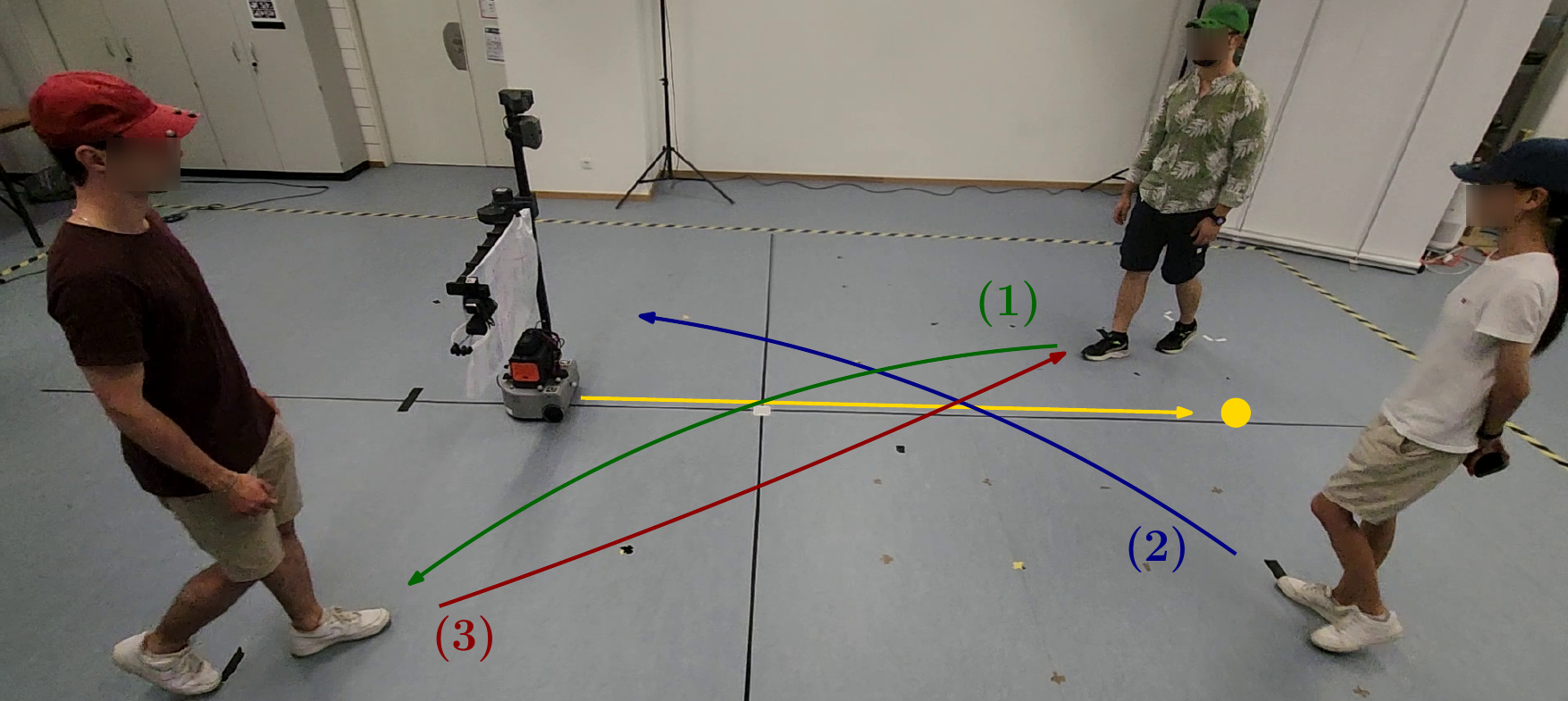}
    \caption{Experimental setup for the banner and cup tasks. Each scenario involves $\nhum \in \{1,2,3\}$ humans instructed to follow the illustrated trajectories. The robot starts at the illustrated position and moves toward the yellow circle, resulting in an interaction between all agents at the center of the room.}
    \label{fig:banner_cup_setup}
    \reducespace
\end{figure}

We evaluate the performance of \ac{\smith} with three sets of experiments:
\textit{(i)} we execute two tasks simultaneously, switching their priorities (Section~\ref{sec:banner_cup}),
\textit{(ii)} we perform a sequential pick-and-place task with different human-motion models on two different platforms (Section~\ref{sec:pick_and_place}), and
\textit{(iii)} we evaluate adversarial interactions in which a human purposefully blocks the motion of the robot.

For all experiments, we implement the \ac{\smith} optimization problem in Python solving the \ac{MPC} in \eqref{eq:smith} using the \texttt{acados} library \cite{verschueren2022acados}.
Human positions are measured by external motion capture systems; however, this framework is agnostic to perception and can be extended to use onboard sensors \cite{samavi2025diffusion}. An exhaustive collection of experiments can be found in the accompanying video: \url{http://tiny.cc/sm2ith}. 

\subsection{Shaping Robot Behavior Through Task Prioritization}\label{sec:banner_cup}

\begin{figure}[t]
	\centering
    \begin{subfigure}{0.49\linewidth}
        \centering
        \includegraphics[width=\textwidth]{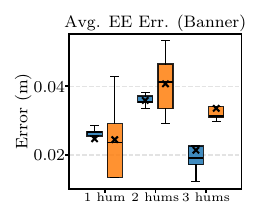}
        \caption{Avg. EE Error}
        \label{subfig:banner_cup_avg_ee_err}
    \end{subfigure}
    \hfill
    \begin{subfigure}{0.49\linewidth}
        \centering
        \includegraphics[width=\textwidth]{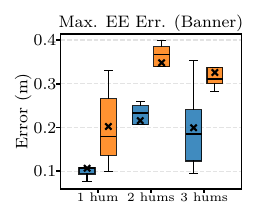}
        \caption{Max. EE Error}
        \label{subfig:banner_cup_max_ee_err}
    \end{subfigure}
    \begin{subfigure}{0.49\linewidth}
        \centering
        \includegraphics[width=\textwidth]{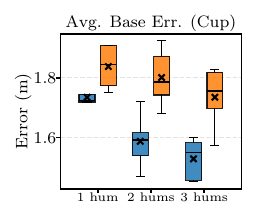}
        \caption{Avg. Base Error}
        \label{subfig:banner_cup_avg_base_err}
    \end{subfigure}
    \hfill
    \begin{subfigure}{0.49\linewidth}
        \centering
        \includegraphics[width=\textwidth]{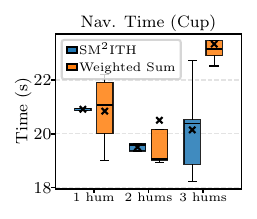}
        \caption{Navigation Time}        \label{subfig:banner_cup_nav_time}
    \end{subfigure}
    \caption{Box-and-whisker plots summarizing the results of task prioritization experiments comparing the \smith{} and the weighted sum method.
    For cases where the end effector task is prioritized we present the average and maximum end-effector tracking error and the navigation efficiency in the Banner tasks (end effector task prioritized E$\ge$B) and in the Cup task (base task prioritized B$\ge$E).
    Median values are illustrated by the line in the box-and-whisker plots and mean values are indicated by the $\times$. At higher human densities (3 humans), the proposed \ac{\smith} method outperforms the weighted sum controller on the high-priority task, highlighting the benefits of strict lexicographic prioritization.}
    \label{fig:banner_cup_results}
    \reducespace
\end{figure}

To evaluate how task priority shapes the behavior of \ac{\smith}, we formulate a scenario where the robot needs to perform two tasks simultaneously while avoiding humans. We perform these experiments with the Hello-Robot Stretch~3, a 7-\ac{DOF} platform with a nonholonomic base and a low-redundancy arm.

\begin{figure}[t]
    \centering
    \includegraphics[width=\columnwidth]{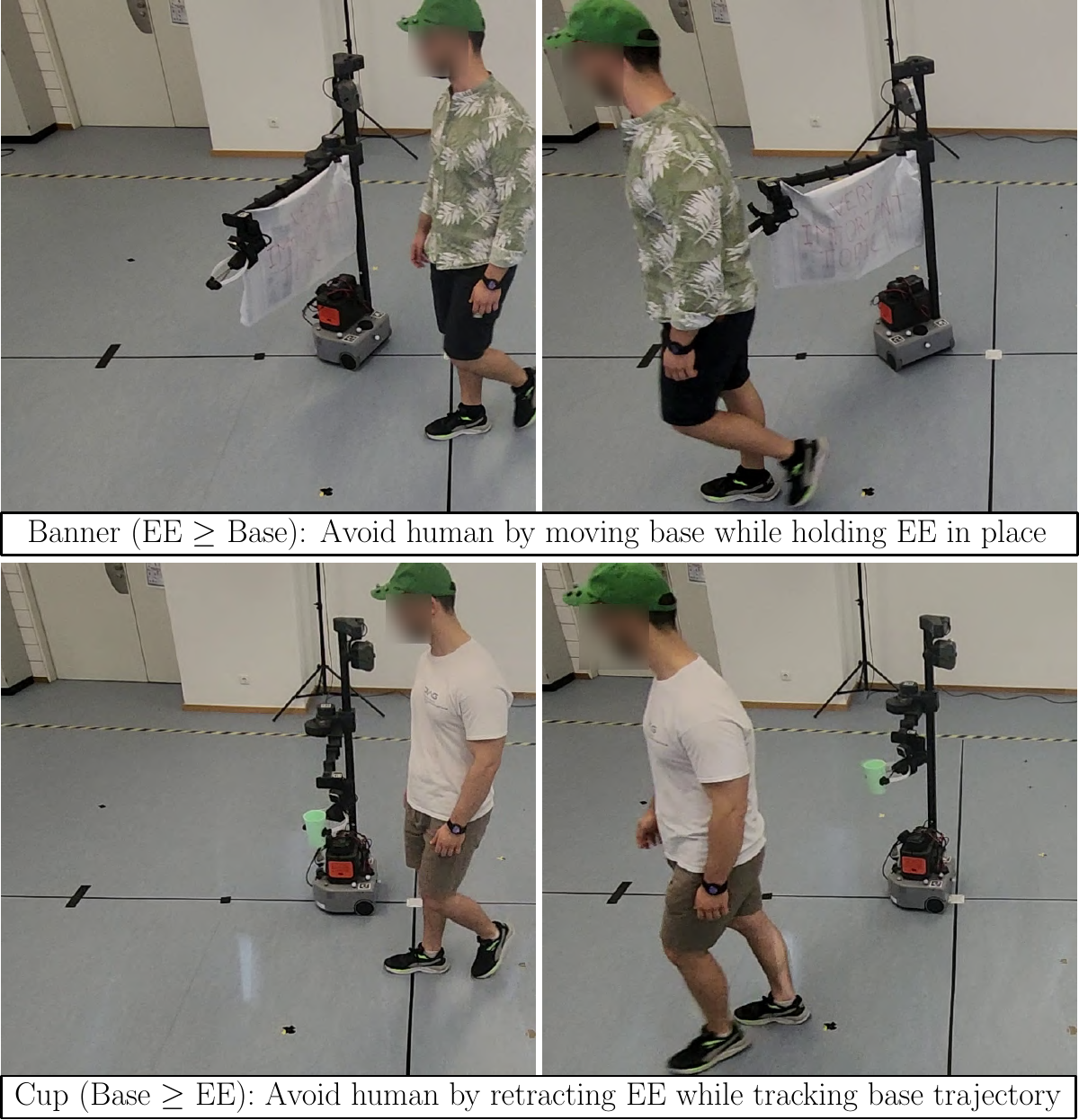}
    \caption{In both scenarios, the human slowly approaches the robot arm, and the robot has two options to avoid collision, either by retracting the end effector or moving the base. In the banner transport scenario (top), the end effector task has higher priority and the base of the robot moves to avoid collision with the human, while in the cup transport scenario (bottom), the base task has higher priority, and the end effector retracts instead. These examples showcase the potential of leveraging hierarchical formulation to encode different robot behaviors based on the context.}
    \label{fig:banner_cup_qualitative}
\end{figure}

\textit{Task Specification}: Fig.~\ref{fig:banner_cup_setup} illustrates the experimental setup. In an environment with $\nhum=\{1,2,3\}$ humans, having the same initial positions, illustrated in Fig.~\ref{fig:banner_cup_setup}, the robot tasks consist of a base navigation task, where the robot is required to reach a predetermined position on the opposite side of the workspace (the yellow circle in Fig.~\ref{fig:banner_cup_setup}), and an \ac{EE} positioning task, where the robot is required to keep its arm extended away from the base.
We design two scenarios with reversed relative priority between the tasks (Fig.~\ref{fig:banner_cup_qualitative}):
\begin{itemize}
    \item EE task has higher priority than base task (E$\geq$B): This task prioritization is motivated by a ``banner-carrying'' task. The main priority of the robot is to display a banner mounted on its arm with a secondary objective of navigating across the room.
    \item Base task has higher priority than EE task (B$\geq$E): This task is motivated by a robot waiter carrying a cup of water. The main priority of the robot is to navigate to its goal position, but as a secondary objective it prefers to keep the cup away from its base to avoid spilling onto itself.
\end{itemize}

\textit{Evaluated Control Methods}: To evaluate the effect of the strict prioritization in \ac{\smith}, we compare our approach with a weighted-sum \ac{MPC} baseline. The baseline is composed of the same costs and constraints as \eqref{eq:smith}, but instead of using the lexicographic framework, it optimizes a single whole-body \ac{MPC} problem combining the two tasks with a weighted sum of costs
\begin{equation*}
    \min_{\mpcstate, \mpccontrol, \mpcaction, \mpcslack, \mpcmul}
    \weight_{\text{ee}}\taskcost_{\text{ee}}
    + \weight_{\text{base}}\taskcost_{\text{base}}
    + \taskcost_{\text{eff}},
\end{equation*}
where $\weight_{ee}$ and $\weight_{base}$ are scalar weights chosen to approximate hierarchical preferences.

\textit{Evaluation Metrics}: To evaluate performance on the \ac{EE} task, we calculate average and maximum end-effector tracking error over each run. To evaluate performance on the base task, we calculate total navigation time per run, and average deviation from the optimal base path over each run.

\textit{Results and Discussion}: We run each experiment scenario five times with each control method for a total of 60 runs. We summarize the quantitative results in Fig.~\ref{fig:banner_cup_results} comparing the \smith{} and the weighted sum approach. When the end-effector task has higher priority (banner experiment), we observe that the average \ac{EE} tracking errors (Fig.~\ref{subfig:banner_cup_avg_ee_err}) of \ac{\smith} are similar to the weighted-sum controller at low human densities ($\nhum=\{1,2\}$), but \ac{\smith} performs significantly better at a high human density ($\nhum=3$). Moreover, if we look at the maximum \ac{EE} tracking errors (Fig.~\ref{subfig:banner_cup_max_ee_err}), we can see that in all human densities the error over the run is lower when using \ac{\smith} than the baseline. Similarly, when the base navigation task has higher priority (cup experiments), we observe that the \ac{\smith} methods completed the tasks more efficiently in terms of both metrics (Figs.~\ref{subfig:banner_cup_avg_base_err}, \ref{subfig:banner_cup_nav_time}).
The weighted-sum baseline performs worse in the high-priority task because it does not enforce a strict task hierarchy. Moreover, the optimal weights that approximate task prioritization are not consistent across scenarios.

\subsection{Human Motion Prediction in Pick-and-Place Tasks} \label{sec:pick_and_place}

\begin{figure}[t]
    \centering
    \includegraphics[width=\linewidth]{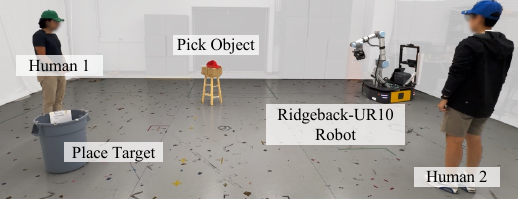}
    \caption{Experimental setup for the pick-and-place experiments. The robot starts from the illustrated position and must pick the red hat and place it into the target and end the run by returning to the start point. Each scenario involves $\nhum \in \{1,2\}$ humans, starting at the illustrated positions and crossing to the opposite side of the room each time the robot travels between start, pick, place, and end locations.}
    \label{fig:pick_and_place_setup}
    \reducespace
\end{figure}
\begin{figure*}[t]
	\centering
    \begin{subfigure}{0.49\linewidth}
		\centering
        \begin{subfigure}{0.49\columnwidth}
		      \includegraphics[width=\textwidth]{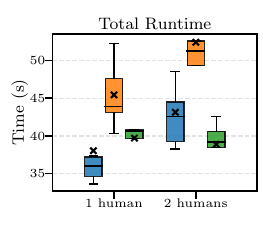}
        \end{subfigure}
        \begin{subfigure}{0.49\columnwidth}
		      \includegraphics[width=\textwidth]{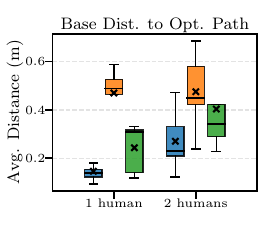}
        \end{subfigure}
		\caption{Ridgeback-UR10 Navigation Efficiency}
        \label{subfig:pick_and_place_thing_navigation}
	\end{subfigure}
    \hfill
	\begin{subfigure}{0.49\linewidth}
		\centering
        \begin{subfigure}{0.49\columnwidth}
		      \includegraphics[width=\textwidth]{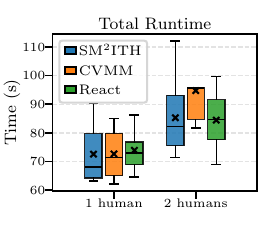}
        \end{subfigure}
        \begin{subfigure}{0.49\columnwidth}
		      \includegraphics[width=\textwidth]{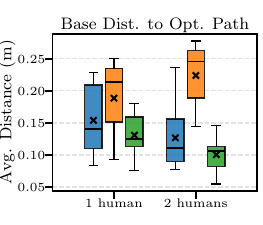}
        \end{subfigure}
		\caption{Stretch 3 Navigation Efficiency}
        \label{subfig:pick_and_place_stretch_navigation}
	\end{subfigure}
    \hfill
    \begin{subfigure}{0.49\linewidth}
        \centering
        \begin{subfigure}{0.49\columnwidth}
		      \includegraphics[width=\textwidth]{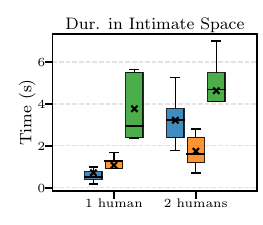}
        \end{subfigure}
        \begin{subfigure}{0.49\columnwidth}
		      \includegraphics[width=\textwidth]{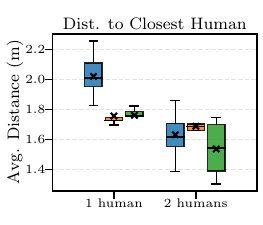}
        \end{subfigure}
        \caption{Ridgeback-UR10 Robot-Human Proxemics}
        \label{subfig:pick_and_place_thing_proxemic}
    \end{subfigure}
    \hfill
    \begin{subfigure}{0.49\linewidth}
        \centering
        \begin{subfigure}{0.49\columnwidth}
		      \includegraphics[width=\textwidth]{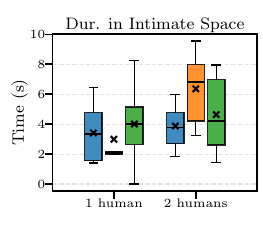}
        \end{subfigure}
        \begin{subfigure}{0.49\columnwidth}
		      \includegraphics[width=\textwidth]{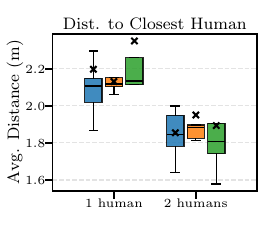}
        \end{subfigure}
        \caption{Stretch 3 Robot-Human Proxemics}
        \label{subfig:pick_and_place_stretch_proxemic}
    \end{subfigure}
    \caption{Box-and-whisker plots summarizing the quantitative results of the pick and place experiments presented in Fig.~\ref{fig:pick_and_place_setup}. We compare \ac{\smith}, with controllers using \ac{CVMM}, and Reactive models on two platforms. Median values are illustrated by the line in the box-and-whisker plots and mean values are indicated by the $\times$. Results show that \ac{CVMM} diverges most from optimal paths, while interactive \ac{\smith} predictions balance safer proxemics with task efficiency.}
    \label{fig:pick_and_place_results}
    \reducespace
\end{figure*}
To evaluate how human models affect the robot's behaviors, we task the robot to pick and place objects at designated positions while performing collision avoidance with humans. To study the impact of different levels or redundancy and kinematic structure, we perform the experiments with both the Hello-Robot Stretch~3, a 7-\ac{DOF} platform with a non-holonomic base, and the Ridgeback-UR10, a 9-\ac{DOF} platform with a holonomic base, illustrated in Fig.~\ref{fig:front_figure}.

\textit{Task Specification}:
Fig. \ref{fig:pick_and_place_setup} illustrates the experimental setup. In an environment with $n_h = \{1, 2\}$ humans, the pick-and-place task consists of a sequence of EE and base targets that lead the robot to pick up the red hat before placing it at a target location.

\textit{Evaluated Control Methods}:
To evaluate the effect of the closed-loop human prediction model integrated into \ac{\smith}, we compare our proposed approach with two baseline human motion models within the same control framework:
\begin{itemize}
    \item SM$^2$ITH: Our proposed method that interactively models human motion by embedding the \ac{ORCA} problem~\eqref{eq:relaxed_orca_acc_argmin} as a lower-level constraint in the controller.
    \item CVMM: A non-interactive predict-then-plan baseline comprised of the same control formulation as SM$^2$ITH except that the interactive human model is replaced with a \ac{CVMM}-based human prediction over the horizon. The predictions are pre-generated before each control step~$\ctime$ and stay the same during the \ac{MPC} optimization.
    \item React: A reactive baseline comprised of the same control formulation as the above two methods, except without any prediction of human motion. This method only reacts to the current tracked positions of the humans.
\end{itemize}

\textit{Evaluation Metrics}:
To evaluate the effects of human models on the robot's whole-body navigation efficiency we measure total runtime and the distance of the base from the optimal path, and to evaluate human-robot proxemics, we measure the duration spent in any human's intimate space (defined as a 0.45 $m$ circle around a human \cite{hall1990hidden}), and average robot–human distance over each run.

\textit{Results and Discussions}:
For each platform, we conduct ten trials for \ac{\smith} and five trials for each of the other two models in an environment considering either one or two humans resulting in 80 total runs.

From the navigation metrics for both robots (Figs. \ref{subfig:pick_and_place_thing_navigation}, \ref{subfig:pick_and_place_stretch_navigation}), we can observe that \ac{CVMM} consistently yields the largest deviations from the optimal path, leading to higher runtimes on both platforms. The reactive formulation remains closer to the optimal path but lacks prediction, frequently cutting across human trajectories and compromising safety, as illustrated in Fig.~\ref{subfig:pick_and_place_thing_proxemic} by the longer duration the Ridgeback-UR10 robot spends in the humans' intimate space. Conversely, \ac{CVMM} reduces time spent in this space but at the cost of degraded navigation efficiency. In contrast, the \ac{\smith} framework balances efficiency and safety by adapting predictions to minimize unnecessary detours while respecting human motion.

Unlike the Ridgeback-UR10, the proxemics results for the Stretch 3 robot (Fig.~\ref{subfig:pick_and_place_stretch_proxemic}) do not demonstrate significant differences between the evaluated methods. We attribute this difference to platform design. In the navigation results (Figs.~\ref{subfig:pick_and_place_thing_navigation},~\ref{subfig:pick_and_place_stretch_navigation}), we can observe that using the Stretch 3 results in all methods having longer runtimes, highlighting the low speed of this robot (maximum 0.3m/s). The Ridgeback–UR10, with its holonomic base and greater redundancy, spends far less time in the intimate space of humans, as it can sidestep approaching agents with agility. On the other hand, the nonholonomic Stretch 3 requires complex maneuvers to orient its base and arm away from humans, resulting in increased close interactions.

We also demonstrate that our proposed framework generalizes to randomized pick-and-place targets on both robots. The robots successfully navigate in the human-centered environment avoiding humans while performing the tasks. The results are shown in the accompanying video.

\subsection{Adversarial Human Interaction}

\begin{figure}
    \centering
    \includegraphics[width=\columnwidth]{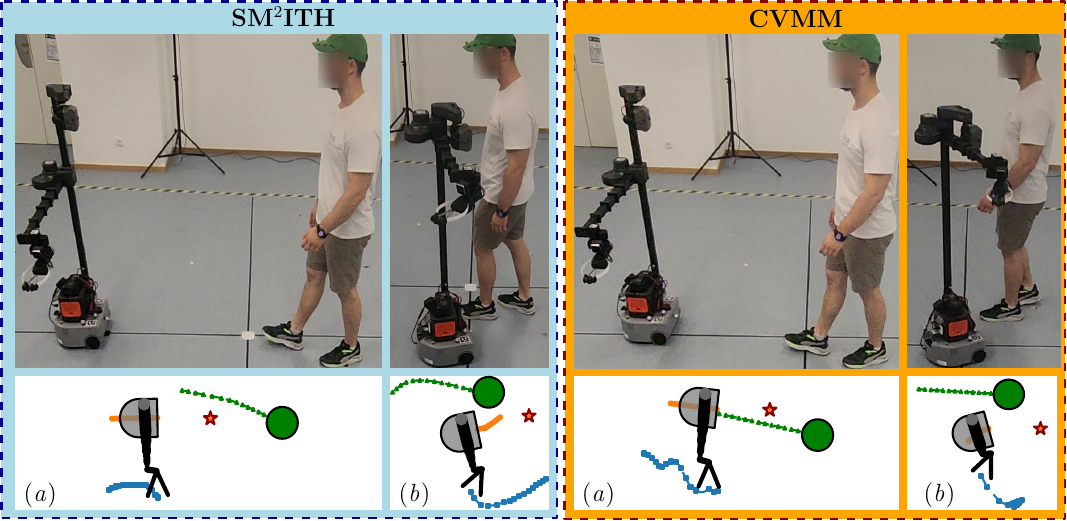}
    \caption{Comparison of adversarial human interaction using the proposed \smith{} (left) and CVMM baseline (right). Each model is shown through experimental snapshots (top row) and reconstructed animations with predicted trajectories (bottom row). The orange star illustrates the robot base target, the orange and blue lines illustrate the base and EE plans, and the green line illustrates the human prediction. Whereas in the first snapshot (a) the robot using either method plans to retreat, in the second snapshot (b) the interactive predictions in \smith{} allows the robot to reason about the coupled human–robot trajectories resulting in a robot plan that continues towards the target.}
    \label{fig:bullying_orca_vs_cvmm}
    \reducespace
\end{figure}

We also qualitatively evaluate the effect of human motion prediction in adversarial situations. We formulate a scenario where the human deliberately walks directly towards the robot, effectively forcing it to retreat. We compare \ac{\smith} with the \ac{CVMM} baseline using the Stretch 3 platform.

We illustrate this scenario in Fig.~\ref{fig:bullying_orca_vs_cvmm}. In the first snapshot for both methods (a), the human is walking towards the robot causing the robot base plan (orange) to back away. In the second snapshot (b) as the human gets closer to the robot, the \ac{\smith} framework generates a joint human prediction (green) and robot plan (orange for base, blue for EE) where both agents proceed while avoiding each other. In contrast, the \ac{CVMM} model extrapolates human velocity in a straight line, resulting in a robot plan that continues retreating. The interactive nature of \ac{\smith} allows the robot to anticipate that the human can adapt to its presence.

\section{Conclusions and Future Works}

In this work, we introduce \ac{\smith}, a unified framework that integrates \ac{HTMPC} with interactive human motion prediction for safe and efficient mobile manipulation in human-centered environments. By combining bilevel optimization with a lexicographic task hierarchy, \ac{\smith} enables robots to jointly reason about manipulation, navigation, and human–robot interactions. Our experiments on two different platforms, the Stretch~3 and the Ridgeback–UR10, demonstrate that \ac{\smith} achieves better task prioritization compared to a weighted-sum baseline and safe human-robot proxemics. We also demonstrate interactive robot behavior during adversarial interactions.

As future work, we are interested in further \textit{(i)} incorporating active perception of both the environment and humans using onboard sensors (e.g., using a 6D pose estimation of human body~\cite{henning2023bodyslam++} to model human anatomy and thereby obtaining even less conservative human collision avoidance constraints), \textit{(ii)} leveraging additional models to capture more complex social behaviors (e.g., group dynamics \cite{moussaid2010walking}, human intent-driven motion \cite{abbate2024self}, or leveraging learning-based generative models in motion prediction \cite{samavi2025diffusion}), and ($iii$) conducting experiments considering more cluttered environments to evaluate the framework's scalability and its ability to maintain efficient navigation within high-density crowds.

\bibliographystyle{IEEEtran}
\bibliography{IEEEabrv, references}

\end{document}

%% file: tools/paper_macros.tex
\newcommand{\dtime}{k}               
\newcommand{\ctime}{t}               
\newcommand{\kpone}{{\dtime+1}}      
\newcommand{\deltat}{{\Delta\ctime}} 

\newcommand{\horiz}{N}          
\newcommand{\nhum}{n_h}         
\newcommand{\nobs}{n_o}         
\newcommand{\ntasks}{L}         
\newcommand{\configdim}{n}      
\newcommand{\pseudoveldim}{m}   

\newcommand{\id}[2]{{#2^{\scriptstyle (#1)}}}   
\newcommand{\robidmarker}{r}                    
\newcommand{\idA}{j}                            
\newcommand{\idB}{l}                            
\newcommand{\idStat}{\tilde{\idB}}              
\newcommand{\rob}[1]{\id{\robidmarker}{#1}}     
\newcommand{\humA}[1]{\id{\idA}{#1}}            
\newcommand{\idtask}{\ell}                      
\newcommand{\idtaskmid}{i}                      
\newcommand{\idtaskdim}{s}                      
\newcommand{\idtaskdimmid}{p}                   

\newcommand{\statescalar}{x}                    
\newcommand{\controlscalar}{u}                  
\newcommand{\actionscalar}{a}                   
\newcommand{\configscalar}{q}                   
\newcommand{\accscalar}{\ddot{\configscalar}}   
\newcommand{\pseudovelscalar}{\nu}              
\newcommand{\state}{\bm\statescalar}            
\newcommand{\control}{\bm\controlscalar}        
\newcommand{\action}{\bm\actionscalar}          
\newcommand{\config}{\bm\configscalar}          
\newcommand{\vel}{\dot\config}                  
\newcommand{\pseudovel}{\bm\pseudovelscalar}    
\newcommand{\pseudoacc}{\dot\pseudovel}         
\newcommand{\at}[2]{{#1}_{\scriptstyle #2}}     
\newcommand{\stateat}[1]{\at{\state}{#1}}       
\newcommand{\workspace}{\wcal}  
\newcommand{\sspace}{\xcal}     
\newcommand{\uspace}{\ucal}     

\newcommand{\robstate}{\rob\state}          
\newcommand{\robconfig}{\rob\config}        
\newcommand{\robvel}{\rob\vel}              
\newcommand{\robpseudovel}{\rob\pseudovel}  
\newcommand{\robpseudoacc}{\rob\pseudoacc}  
\newcommand{\robstateat}[1]{\rob{\state_{#1}}}              
\newcommand{\robconfigat}[1]{\rob{\at{\config}{#1}}}        
\newcommand{\nullkinmat}{\bm G} 
\newcommand{\nullkinmatatstate}{\nullkinmat(\robconfig)} 
\newcommand{\robdyn}{\bm f}                 

\newcommand{\humdyn}{\bm g}                 
\newcommand{\forhumin}{\idA \in \squiggly{1,\ldots, \nhum}}                   
\newcommand{\forobsin}{\idStat \in \squiggly{\nhum + 1,\ldots, \nhum+\nobs}}  

\newcommand{\mpcstate}{\bar{\state}}                        
\newcommand{\mpccontrol}{\bar{\control}}                    
\newcommand{\mpcaction}{\bar{\action}}                      
\newcommand{\mpcslack}{\bar{\orcaslack}}                    
\newcommand{\mpcmul}{\bar{\lagmulvec}}                      
\newcommand{\mpcstateat}[1]{\at{\mpcstate}{#1}}             
\newcommand{\mpcrobstate}{\rob\mpcstate}                    
\newcommand{\mpcrobcontrol}{\mpccontrol}                    
\newcommand{\mpchumslack}{\humA\mpcslack}                   
\newcommand{\mpchummul}{\humA\mpcmul}                       
\newcommand{\mpcrobstateat}[1]{\rob{\at{\mpcstate}{#1}}}    
\newcommand{\mpcrobcontrolat}[1]{\at{\mpccontrol}{#1}}      
\newcommand{\mpchumstateat}[1]{\humA{\at{\mpcstate}{#1}}}   
\newcommand{\mpchumactionat}[1]{\humA{\at{\mpcaction}{#1}}} 
\newcommand{\init}{\text{o}}            
\newcommand{\dist}{d}                   
\newcommand{\taskcost}{\jcal}           
\newcommand{\weight}{W}                 
\newcommand{\costweight}{\bm\weight}    
\newcommand{\lagmul}{\lambda}           
\newcommand{\lagmulvec}{\bm\lagmul}     
\newcommand{\lagfun}{\lcal}             
\newcommand{\errscalar}{e}                  
\newcommand{\err}{\bm \errscalar}           
\newcommand{\scdist}{\bm{\dist}_{\text{sc}}}
\newcommand{\safeset}{\ccal}                
\newcommand{\cbffun}{h}                     
\newcommand{\cbfparam}{\gamma}              
\newcommand{\cbfvec}{\bm\cbffun}            
\newcommand{\orcacontrolscalar}{z}                  
\newcommand{\orcacontrol}{\bm{\orcacontrolscalar}}  
\newcommand{\orcalinedir}{\bm n}                    
\newcommand{\orcaconstgeq}{\rho}                    
\newcommand{\orcaconst}{\bm c}                      
\newcommand{\orcarlxsolnset}{\ocal}                 
\newcommand{\orcapref}{\text{des}}                  
\newcommand{\orcaslackpenal}{\bm M}                     
\newcommand{\orcaslackvar}{\zeta}                   
\newcommand{\orcaslackvel}{\orcaslackvar_v}         
\newcommand{\orcaslackacc}{\orcaslackvar_a}         
\newcommand{\orcaslack}{\bm\orcaslackvar}           
\newcommand{\orcasupcolldyn}{\id{\idA|\idB}}        
\newcommand{\orcasupcollstatic}{\id{\idA|\idStat}}  

\newcommand{\ccal}{\mathcal{C}}

\newcommand{\jcal}{\mathcal{J}}

\newcommand{\lcal}{\mathcal{L}}

\newcommand{\ocal}{\mathcal{O}}

\newcommand{\ucal}{\mathcal{U}}

\newcommand{\wcal}{\mathcal{W}}
\newcommand{\xcal}{\mathcal{X}}

\newcommand{\Reals}{\mathbb{R}}

\newcommand{\squiggly}[1]{\left\{#1\right\}}

\newcommand*\centremathcell[1]{\omit\hfil$\displaystyle#1$\hfil\ignorespaces}

\newcommand{\suchthat}{:}

\newcommand{\trans}[1]{#1^T}

\newcommand{\norm}[1]{\left\lVert#1\right\rVert}
\newcommand{\pnorm}[2]{\norm{#2}_{#1}}
\newcommand{\twonorm}[1]{\pnorm{2}{#1}}

\newcommand{\grad}[2]{\nabla_{\scriptstyle #1} #2}

\DeclareMathOperator*{\argmin}{arg\,min}
\DeclareMathOperator*{\lexmin}{lex\,min}

%% file: tools/acronyms.tex
\acrodef{SM$^2$ITH}{Safe Mobile Manipulation with Interactive Human Prediction via Task-Hierarchical Bilevel Model Predictive Control}
\acrodef{QP}{Quadratic Programming}
\acrodef{MPC}{Model Predictive Control}
\acrodef{NMPC}{Nonlinear Model Predictive Control}
\acrodef{HTMPC}{Hierarchical Task Model Predictive Control}
\acrodef{STMPC}{Single Task Model Predictive Control}
\acrodef{QCQP}{Quadratically Constrained Quadratic Program}
\acrodef{KKT}{Karush-Kuhn-Tucker}
\acrodef{RVO}{Reciprocal Velocity Obstacle}
\acrodef{CBF}{Control Barrier Function}
\acrodef{DT-CBF}{Discrete-Time Control Barrier Function}
\acrodef{DOF}{degrees of freedom}
\acrodef{EE}{end-effector}
\acrodef{CVMM}{Constant Velocity Motion Model}

\newcommand{\smith}{SM$^2$ITH}

\acrodef{MID}{Motion Indeterminacy Diffusion}
\acrodef{T++}{Trajectron++}
\acrodef{JMID}{Joint Motion Indeterminacy Diffusion}
\acrodef{iMID}{Individual Motion Indeterminacy Diffusion}
\acrodef{DDIM}{Denoising Diffusion Implicit Model}
\acrodef{DDPM}{Denoising Diffusion Probabilistic Model}
\acrodef{GRU}{Gated Recurrent Unit}
\acrodefplural{GRU}{Gated Recurrent Units}

\acrodef{AV}{Autonomous vehicle}
\acrodefplural{AVs}{Autonomous vehicles}

\acrodef{LP}{Linear Program}
\acrodef{QCQP}{Quadratically Constrained Quadratic Program}
\acrodef{MPC}{Model Predictive Control}
\acrodef{MPCC}{Mathematical Program with Complementarity Constraints}

\acrodef{KKT}{Karush-Kuhn-Tucker}
\acrodef{LICQ}{Linear Independence Constraint Qualification}
\acrodef{CRCQ}{Constant Rank Constraint Qualification}
\acrodef{SCQ}{Slater's Constraint Qualification}

\acrodef{OGM}{Occupancy Grid Map}
\acrodef{NF}{Navigation Function}
\acrodef{RVO}{Reciprocal Velocity Obstacle}
\acrodefplural{RVO}{Reciprocal Velocity Obstacles}
\acrodef{SICNav}{Safe and Interactive Crowd Navigation}
\acrodef{SICNav-Diffusion}{Safe and Interactive Crowd Navigation with Diffusion Trajectory Predictions}

\acrodef{CAMPC}{Collision Avoiding Model Predictive Control}
\acrodef{RHC}{Receding Horizon Control}
\acrodef{CLF}{control-Lyapunov Function}
\acrodef{DWA}{Dynamic Window Approach}
\acrodef{SFM}{Social Force Model}
\acrodef{ESFM}{Extended Social Force Model}

\acrodef{ORCA}{Optimal Reciprocal Collision Avoidance}
\acrodef{VO}{Velocity Obstacle}
\acrodef{CA}{Collision Avoiding}

\acrodef{ADE}{Average Displacement Error}
\acrodef{FDE}{Final Displacement Error}
\acrodef{SADE}{Scene-level Average Displacement Error}
\acrodef{SFDE}{Scene-level Final Displacement Error}
\acrodef{KDE}{Kernel Density Estimate}
\acrodefplural{KDE}{Kernel Density Estimates}
\acrodef{NLL}{Negative Log Likelihood}

\acrodef{KF}{Kalman Filter}
\acrodef{EKF}{Extended Kalman Filter}
\acrodef{UKF}{Unscented Kalman Filter}
\acrodef{PF}{Particle Filter}
\acrodef{SLAM}{Simultaneous Localization and Mapping}